\pdfoutput=1

\documentclass{article}

\usepackage{acl}
\usepackage{multirow}
\usepackage{times}
\usepackage{latexsym}
\usepackage{amsmath}
\usepackage{amssymb}
\usepackage[T1]{fontenc}

\usepackage[utf8]{inputenc}
\usepackage{color}

\usepackage{microtype}

\usepackage{inconsolata}

\usepackage{graphicx}

\title{Stance-Driven Multimodal Controlled Statement Generation: New Task and Dataset}

\author{
 \textbf{Bingqian Wang\textsuperscript{1}},
 \textbf{Quan Fang\textsuperscript{1}},
 \textbf{Jiachen Sun\textsuperscript{1}},
 \textbf{Xiaoxiao Ma\textsuperscript{1}},
\\
\\
 \textsuperscript{1}Beijing University of Posts and Telecommunications
\\
 \small{
   \textbf{Correspondence:} \href{mailto:email@domain}{qfang@bupt.edu.cn}
 }
}

\begin{document}
\maketitle
\begin{abstract}
Formulating statements that support diverse or controversial stances on specific topics is vital for platforms that enable user expression, reshape political discourse, and drive social critique and information dissemination. With the rise of Large Language Models (LLMs), controllable text generation towards specific stances has become a promising research area with applications in shaping public opinion and commercial marketing. 
However, current datasets often focus solely on pure texts, lacking multimodal content and effective context, particularly in the context of stance detection. 
In this paper, we formally define and study the new problem of stance-driven controllable content generation for tweets with text and images, where given a multimodal post (text and image/video), a model generates a stance-controlled response. To this end, we create the Multimodal Stance Generation Dataset (StanceGen2024),  the first resource explicitly designed for multimodal stance-controllable text generation in political discourse. It includes posts and user comments from the 2024 U.S. presidential election, featuring text, images, videos, and stance annotations to explore how multimodal political content shapes stance expression.                                          
Furthermore, we propose a ~\textbf{~\underline{S}}tance-~\textbf{~\underline{D}}riven ~\textbf{~\underline{M}}ultimodal ~\textbf{~\underline{G}}eneration~\textbf{(SDMG)} framework  that integrates weighted fusion of multimodal features and stance guidance to improve semantic consistency and stance control.
We release the dataset and code\footnote{https://anonymous.4open.science/r/StanceGen-BE9D} for public use and further research.



\end{abstract}

\section{Introduction}
\begin{figure}[t]
\centering
\includegraphics[width=0.5\textwidth]{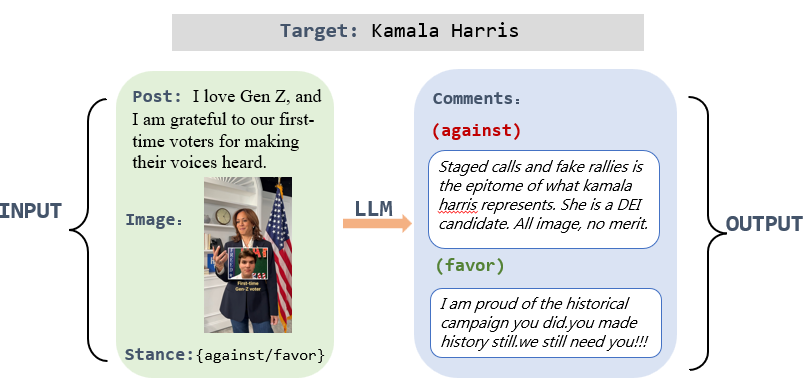} 
\caption{ An overview of our task. The input consists of tweet text, visual images, and a specified stance.}
\label{fig1}
\end{figure}

In the contemporary era of digital interconnectedness, online platforms have emerged as pivotal arenas for political discourse, social critique, and information dissemination. The ability to identify and craft statements that encapsulate the multifaceted, and often divergent, perspectives on specific issues is of paramount importance. Such capability not only empowers users to articulate their viewpoints with greater efficacy but also propels the dynamic evolution of these digital ecosystems. With the advent of generative artificial intelligence (AI) systems built upon large language models (LLMs), automated generating controllable content for a given stance or topic has emerged as a burgeoning research frontier~\cite{DBLP:conf/naacl/SchillerDG21,li2024llmsspeakdiversepeople}, offering the potential to automatically generate texts that consistently align with predetermined stance parameters and other attribute constraints.

While existing studies predominantly focus on textual stance detection~\cite{DBLP:journals/csur/KucukC20,DBLP:journals/corr/abs-2409-15690} which involves classifying textual inputs into discrete categories such as support, opposition, or neutrality. However, the emerging paradigm of generating stance-aligned responses from multimodal inputs - termed \emph{Stance-Driven Multimodal Controlled Statement Generation, SDMCSG} - remains critically underexplored. The aim of SDMCSG is to generate the corresponding statement for a given stance towards a target, which can be an entity, concept, event, idea, opinion, claim, or topic that is either explicitly mentioned or implied within the multimodal input contexts. As illustrated in Figure 1 with a 2024 U.S. presidential campaign example, When presented with Vice President Kamala Harris's supportive stance, as well as her multimodal post featuring campaign text and an official portrait, our framework enables models to generate supportive user comments that maintain ideological consistency with both the visual and textual cues. This capability addresses a critical gap in political communication systems, where authentic opinion expression requires synchronized understanding of multimodal stance indicators and controlled generation of positionally coherent responses.


In order to push forward the research of multimodal stance-driven controlled content generation, we create the Multimodal Stance Generation Dataset (StanceGen2024), the first resource explicitly designed for multimodal stance-controllable text generation in political discourse. This dataset includes posts from candidates and user comments from various social platforms during the 2024 U.S. presidential election, featuring rich text, images, and video content, along with stance annotations. The primary goal of this dataset is to explore how multimodal political content interacts across different media and influences users' stance expression, thereby providing a real and diverse foundation for future multimodal stance generation tasks.
StanceGen2024 is not limited to traditional text data; it also includes multimodal information such as images and videos related to the election, offering more comprehensive contextual information than single-modal text data. These multimodal elements play an essential background role in specific political topics, deepening the semantic connection between textual and visual content. With this data, we aim to explore how to combine text and visual content in the political domain to generate more precise and stance-consistent responses.

Furthermore, we propose an innovative stance-driven multimodal generation framework that optimizes generation effects by weighted fusion of multimodal features and stance guidance. In this framework, we not only consider the varying importance of modalities such as text and images but also apply weighted processing to the features of each modality, ensuring that the generated text maintains semantic consistency while better adhering to the stance requirements. Through this fusion strategy, we can effectively enhance the fluency, relevance, and stance control of the generated content, making the text more aligned with user expectations and accurately reflecting the diversity and complexity of political discourse. Based on this, we improved and fine-tuned the LLaVa open-source model with instruction-based tuning. The results show that our approach achieves a balance between controllability and generation quality, yielding favorable outcomes.

Our main contributions are as follows:
\begin{itemize}
    \item  We introduce StanceGen2024, the first multimodal dataset explicitly designed for stance-controlled generation in political discourse. It pairs multimodal posts (text, images, videos) from the 2024 U.S. presidential election with stance-annotated user responses, enabling systematic exploration of how multimodal context shapes ideological expression.
    \item We propose a novel framework integrating weighted cross-modal attention and stance guidance mechanisms. This architecture dynamically prioritizes stance-critical features (e.g., politically charged visuals) and enforces stance consistency during generation, addressing the limitations of text-centric approaches.
    \item  A series of experiments on our datasets demonstrate that our method is effective and provides a new insight.    
\end{itemize}


\section{Related Work}
\subsection{Related Datasets for Stance-Controlled Generation}
Currently, there is no specialized dataset designed for the generation of text controlled by stance. Traditional controllable text generation tasks~\cite{liang2024controlled,liu2024multi} have utilized sentiment-focused datasets such as the SST-5 dataset~\cite{socher-etal-2013-recursive} and IMDB~\cite{maas-etal-2011-learning}. Two popular datasets like P-Stance~\cite{li2021p} and Twitter Stance Election 2020~\cite{liang2024multi}, are used for stance detection tasks. The P-Stance dataset is a large-scale stance detection resource, consisting of 21,574 tweets extracted from over 2.8 million tweets collected from Twitter, and it only contains pure text. Twitter Stance Election 2020 is a multimodal stance detection dataset used for detecting stances in multimodal content. Both of these datasets are collected using specific labels. To the best of our knowledge, there exists no publicly available dataset that supports stance-controlled statement generation with both multimodal integration and contextual interaction capabilities. Our work addresses this critical limitation by introducing StanceGen2024, a novel benchmark that combines target topic with multimodal features in political discourse.


\subsection{Controllable Text Generation}
LLMs have introduced new methods for controllable text generation, enhancing the manipulation of text attributes. Post-processing techniques~\cite{yang-klein-2021-fudge} allow modifications after generation to control attributes, while prefix tuning~\cite{DBLP:conf/acl/Qian0SWC22} adjusts the initial prompts to guide the generation process. Aspect-controlled content generation has also gained attention, with early work~\cite{DBLP:conf/naacl/SchillerDG21} enabling control over topics, stances, and aspects at the sentence level. Recent advancements in stance-driven text generation include the PCTG-X model~\cite{yang2024topic}, DEBATUNE~\cite{li2024llmsspeakdiversepeople}, and DATG~\cite{liang2024controlledtextgenerationlarge}. These methods primarily focus on text-based datasets and do not address how to handle ultimodal inputs effectively.
We propose a novel stance-driven multimodal controlled statement generation framework that integrates weighted fusion of multimodal features and stance guidance to improve semantic consistency and stance controllability. 

\section{Building the Dataset}

This section details the creation and specifics of the Multimodal Stance Generation Dataset (StanceGen2024). StanceGen2024 is a novel dataset designed for multi-modal stance-controllable text generation, focusing on political discourse during the 2024 U.S. Presidential Election.   The dataset comprises posts from the official Twitter profiles of Kamala Harris and Donald Trump, along with user comments. 
Unlike existing datasets~\cite{li2021p,liang2024multi}, which primarily support stance detection and are often limited to textual content, MTSE2024 is designed to facilitate stance-controlled text generation with rich multi-modal information. While previous multi-modal datasets mostly rely on tweets collected through specific hashtags, they often lack an explicit connection between posts and responses.   In contrast, StanceGen2024 explicitly captures the interaction between tweets and their corresponding comments.  This provides a more realistic training resource for studying context-aware stance generation.

\subsection{Data Construction}
We use the Twitter Streaming API to collect tweets. Similar to previous works ~\cite{mohammad-etal-2016-semeval,conforti-etal-2020-will} that focused on presidential candidates, we concentrate on two political figures in the 2024 presidential election: Donald Trump and Kamala Harris. The collection period spans from July 21, 2024, when Harris replaced Biden as the Democratic presidential candidate, to November 6, 2024, when the election results were announced. We directly collect posts from the two candidates' Twitter profiles during this period, along with user comments under these posts. 
For both posts and user comments, we retain English text and tweets that contain at least one image or a video/GIF. For videos and GIFs, we keep only their first frame, as consecutive frames often contain highly similar visual information. For posts with multiple images, we pair each image with the corresponding text to form multiple samples. 


Given the complexity of the stance-driven multimodal controlled statement generation task, considerable effort must be dedicated to ensuring the dataset's quality, effectiveness, and comprehensiveness. Our focus is on the following key aspects:

\textbf{(1) Multimodal Unified Timestamps:} We synchronized timestamps across text, images, and videos to ensure the correct alignment of different data modalities.

\textbf{(2) Annotation Quality Control:} Annotators underwent training, which included a review of the context surrounding candidates' posts and relevant news during the 2024 campaign.   Before starting, annotators had to pass a preliminary test to ensure their understanding of the task and the nuances of the political context.

\textbf{(3) Topic Segmentation:} We categorized the posts into broad themes based on their political content, such as appeals for support, policy discussion, and campaign highlights, providing a structured overview of the election discourse.

Beyond stance-controlled text generation, StanceGen2024 is also well-suited for a variety of other tasks, including multimodal stance detection, political discourse analysis, and sentiment analysis.  This versatility makes the dataset a valuable resource for understanding political communication and generating contextually aligned responses.

\subsection{Preprocessing}
To ensure dataset quality, we applied several preprocessing steps: 
\textbf{1)} We retained tweets with 10 to 128 words, excluding those outside this range to balance informativeness and conciseness. 
\textbf{2)} We removed irrelevant content, including URLs, @usernames, and unnecessary punctuation, while preserving functional punctuation and meaningful emojis or special characters. 
\textbf{3)} Only English tweets were kept to focus on building an English stance-controllable dataset.

\subsection{Data Annotation}

Our dataset is centered on multimodal stance generation within the context of political discourse. We meticulously annotate both tweets and their associated user comments with political stances (e.g., against or favor) and with topic categories that capture broad themes such as voter mobilization, political ideology, and candidate image projection.   Given the complexity of integrating textual and visual modalities, our annotation process is executed in two stages.

In the initial stage, due to the widely recognized text comprehension capabilities of large-scale models, we employ several large-scale models (GPT-4o~\cite{DBLP:journals/corr/abs-2410-21276}, DeepSeek-V3~\cite{DBLP:journals/corr/abs-2412-19437} and Qwen 2.5-Max~\cite{qwen25}) to perform coarse-grained annotations of stances and topics.   For instances where model outputs are highly consistent, the stance is considered clear;   however, for cases with inconsistent annotations—which may indicate ambiguity or neutrality—manual fine-grained calibration is conducted.   To this end, we engaged three graduate students specializing in multimodal research to serve as annotators for this calibration process.   These annotators received comprehensive training covering key political events during the 2024 campaign, the context behind the candidates’ posts, and guidelines for interpreting multimodal content.  Only those who successfully passed a rigorous preliminary test were permitted to proceed with the formal annotation.

To ensure consistency and reliability, each data instance was independently annotated by two annotators.   In cases of disagreement, a third annotator reviewed the sample and determined the final label.   This meticulous process not only guarantees a high standard of annotation quality but also renders the dataset a valuable resource for a range of applications beyond stance generation, including multimodal stance detection and political discourse analysis.

\subsection{Quality Assessment}
We evaluate inter-annotator agreement using Cohen’s Kappa Statistic~\cite{cohen1960coefficient}, with an average score of 0.719 for StanceGen2024. This indicates substantial agreement between annotators. Additionally, Cohen’s Kappa in related stance detection datasets~\cite{liang2024multi} typically hovers around 0.7, further validating the high quality of our dataset.

\section{Dataset Characteristics}
\begin{table}[!t]
\centering
\scalebox{0.71}{
\begin{tabular}{c|c|c|c|c|c}
\hline
\textbf{Candidate} & \textbf{Posts} & \textbf{Post Images} & \textbf{Favor} & \textbf{Against}  & \textbf{Samples}   \\ 
\hline
Harris            & 837       & 199               & 1,596          
& 10,529        &    12,126        \\ \hline
Trump             & 202     & 156                 & 5,269           & 7,630     &   12,899 \\ \hline
\end{tabular}
}
\caption{Statistics of the StanceGen2024 Dataset.}
\label{tab:posts_comments}
\end{table}
Our multimodal dataset consists of 1,039 posts and 25,025 comments, primarily focusing on political discourse during the 2024 U.S. presidential election, as detailed in Table 1.

\begin{figure}[!t]
\centering
\includegraphics[width=0.5\textwidth]{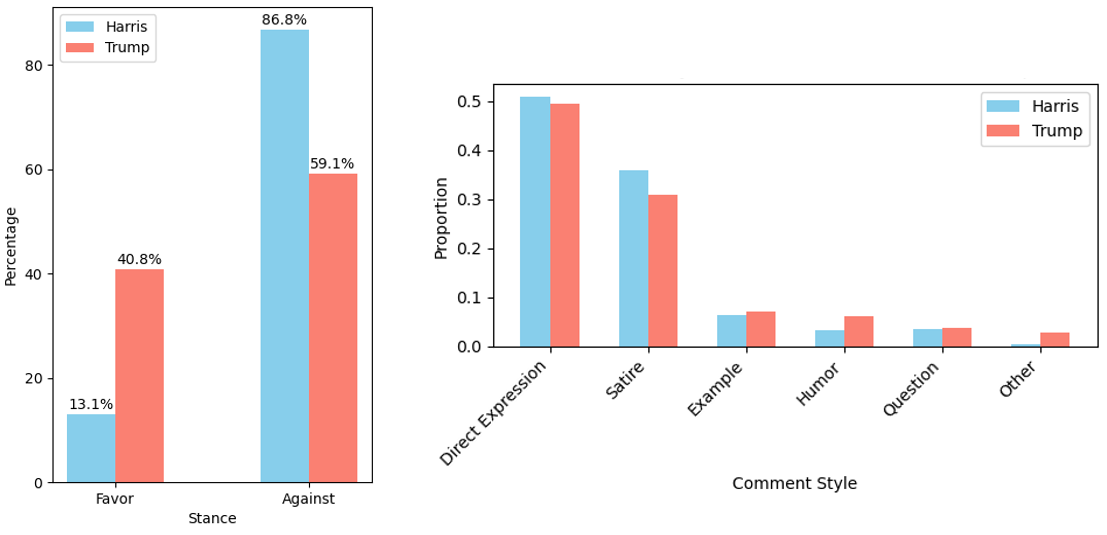} 
\vspace{1mm}
\caption{Comparison of Comment Categories for Harris and Trump}
\label{fig2}
\end{figure}

Through an analysis of the post content, we categorize them into four main types: \textbf{Calls for Voter Support}, \textbf{Sharing Political Ideologies}, \textbf{Self-Promotionand} and \textbf{Reporting Achievements}, with a category labeled as "Other" for posts that do not clearly fall into these categories. These classifications reflect the key topics of discourse shared by the candidates on Twitter and illustrate the different ways they interacted with voters via social media during the election period. The specific distribution is shown in Figure~\ref{fig2}.

\begin{figure}[!t]
\centering
\includegraphics[width=0.4\textwidth]{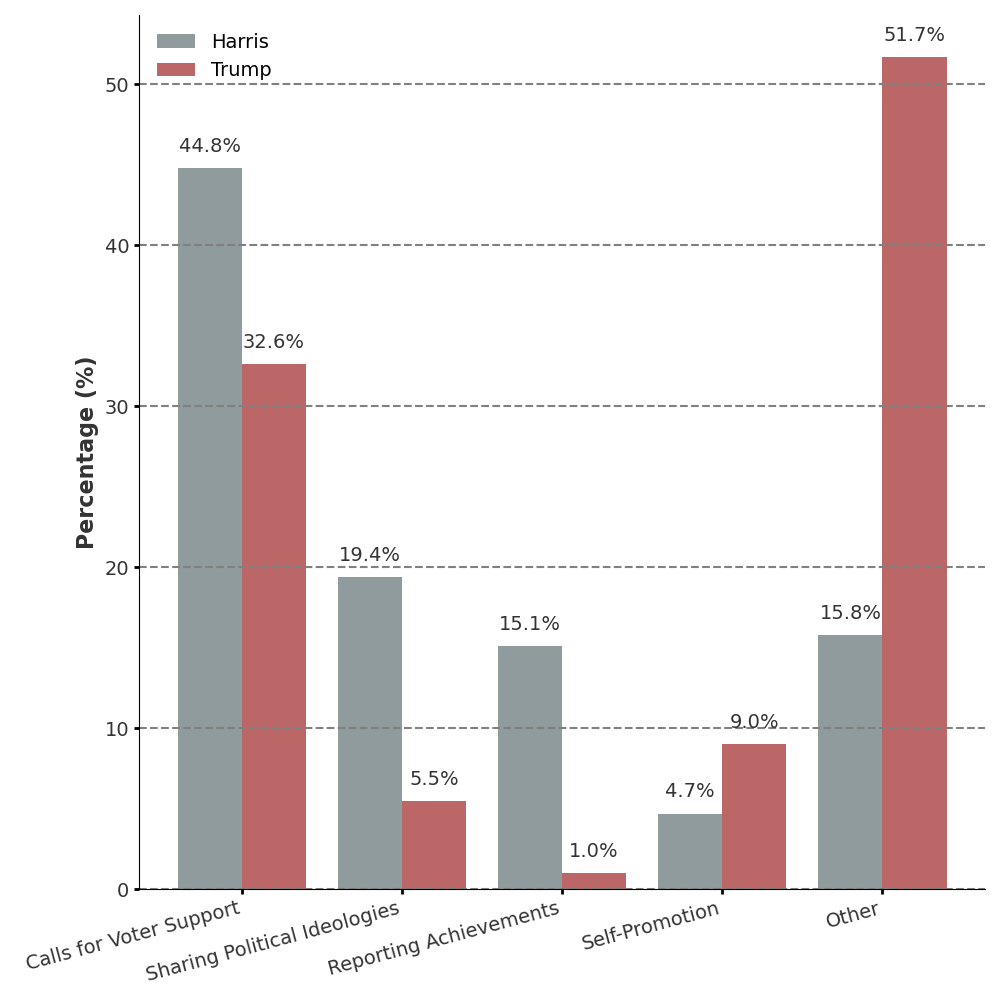}
\caption{Comparison of Post Categories between Harris and Trump}
\label{fig3}
\end{figure}
Regarding the comments, we annotated both the \textbf{stance} (support or opposition) and the \textbf{comment style} for each entry. The comment styles are mainly divided into \textbf{Sarcasm}, \textbf{Direct Expression}, \textbf{Examples}, \textbf{Questions/Counterquestions}, \textbf{Humor/Irony}, and other categories. These styles demonstrate the different ways users express their attitudes toward the candidates and their posts. For Harris’s posts, 86.8\% of the comments were oppositional, while only 13.1\% expressed support. For Trump’s posts, 59.1\% of comments were oppositional, and 40.8\% expressed support. These figures align with the public sentiment during the election period and the eventual election outcome, indicating a higher level of opposition to Harris. The  distribution is shown in Figure~\ref{fig3}.

Given the multimodal nature of the dataset, we also analyzed the proportion of comments that included visual content. Since user comments do not always include images or videos, some comments are purely textual. In the final dataset, 26.6\% of the comments included images, while 8.9\% included videos. This distribution shows that while most comments are text-based, multimodal elements still play a role in enriching the expression of comments and advancing multimodal stance generation.

\section{Methodology}

\begin{figure*}[t]
\centering
\includegraphics[width=0.9\textwidth]{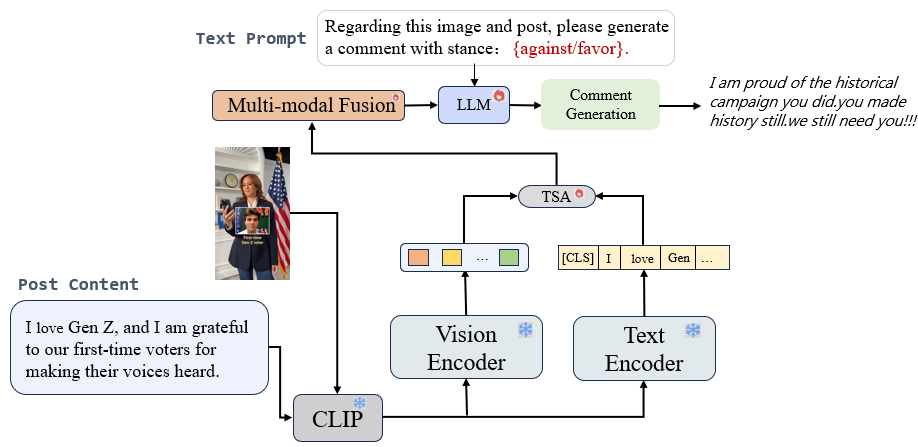}
\vspace{-1mm}
\caption{The overall architecture of our proposed method SDMG.}
\label{fig4}
\end{figure*}
In this section, we will introduce in detail our proposed Stance-driven Multimodal Generation (SDMG) Framework. Given a text $S$, an image $I$ and a specific stance $y$, the goal of multi-modal stance-controlled text generation is to generate a response $R$ that aligns with a specific stance label $y$ for a target $t$, based on $S$ and $I$. To achieve this, we propose a stance-driven multimodal generation framework that leverages both textual and visual modalities. Our framework integrates a weighted fusion of multimodal features and stance guidance, prompting pre-trained models to generate contextually consistent and stance-controlled responses. The architecture of our proposed framework is illustrated in Figure~\ref{fig4}.

\subsection{Visual Encoder}
We adopt the Vision Transformer (ViT) architecture based on the CLIP model~\cite{radford2021learning} to process image information. ViT splits the input image into an \( N \times N \) sequence of image patches and utilizes the Transformer structure to extract image features. On this basis, we introduce a learnable target prompt vector \( P_V \) and insert it into the ViT input sequence, thus guiding the model to focus on specific target areas (such as people or objects). 

Given the input image \( V_0 \), we first split it into an \( N \times N \) sequence of image patches, which are used as the input to the ViT. The target prompt vector \( P_V \) is introduced as a learnable parameter to help the model focus on specific targets within the image. The input sequence to ViT can be represented as:
\begin{equation}
   X_{\text{input}} = [x_V[\text{CLS}]_0, P_V, V_0]
\end{equation}
where \( x_V[\text{CLS}]_0 \) is the [CLS] token of the first layer, used to aggregate global visual information, \( P_V \) is the target prompt vector guiding the model to focus on specific targets, and \( V_0 \) is the sequence of image patches after splitting.

After processing through ViT, the output of the first layer is:
\begin{equation}
   L_1[x_V[\text{CLS}]_1, Z_1, V_1]
\end{equation}
where \( x_V[\text{CLS}]_k \) is the [CLS] token of the \( k \)-th layer, responsible for aggregating visual information, \( Z_1 \) is the intermediate feature representation from the first layer of the Transformer, and \( V_1 \) is the feature representation of the image patches after the first layer's processing.

\subsection{Textual Encoder}
We adopt the text encoder from the CLIP~\cite{radford2021learning} model. After processing through multiple layers of self-attention mechanisms, the output feature of the text encoder is the embedding of the first [CLS] token \( T \in \mathbb{R}^{d_t} \), which represents the global semantic information of the entire text:
\begin{equation}
    T = \text{Transformer}(T_{\text{input}})_{\text{CLS}}
\end{equation}
where \( \text{Transformer}(T_{\text{input}}) \) represents the text sequence processed by the Transformer network, and the embedding \( T \) of the [CLS] token serves as the global semantic representation of the text.

\subsection{TSA and Multi-modal Fusion}
Building on the textual and visual embeddings, we introduce the Task-Sensitive Attention (TSA) mechanism, which dynamically computes the interaction weights between the visual and textual features to capture task-relevant dependencies. Specifically, TSA utilizes cross-modal attention to model the relationships between visual and textual modalities, ensuring that both contribute effectively to the final output.

\subsubsection{Input Features}

The input features for TSA include the visual feature \( V \in \mathbb{R}^{d_v} \), extracted from the [CLS] token embedding of the visual encoder, and the textual feature \( T \in \mathbb{R}^{d_t} \), extracted from the [CLS] token embedding of the text encoder. These features represent the global semantic information from both the visual and textual modalities, where \( d_v \) and \( d_t \) denote the dimensionalities of the visual and textual features, respectively.

\subsubsection{Feature Projection}

To facilitate attention weight computation, both the visual and textual features are projected into the same dimensional space \( d \). This projection is achieved by using learnable weight matrices:
\begin{equation}
Q = W_q V, \quad K = W_k T, \quad V_f = W_v V
\end{equation}
where \( W_q \in \mathbb{R}^{d \times d_v} \), \( W_k \in \mathbb{R}^{d \times d_t} \), and \( W_v \in \mathbb{R}^{d \times d_v} \) are the weight matrices, and \( Q \in \mathbb{R}^d \), \( K \in \mathbb{R}^d \), and \( V_f \in \mathbb{R}^d \) are the query, key, and value vectors, respectively.

\subsubsection{Attention Weight Calculation}

The attention weight is computed by taking the dot product of the query \( Q \) and key \( K \), followed by normalization using the Softmax function. This yields the attention weights, which are then used to weigh the visual features:
\begin{equation}
\text{Attention}(Q, K, V_f) = \text{Softmax}\left(\frac{Q K^T}{\sqrt{d}}\right) V_f
\end{equation}
where \( Q K^T / \sqrt{d} \) represents the scaled dot-product attention, and Softmax converts the similarity scores into a probability distribution, indicating the importance of textual features for visual features. The final output is the weighted visual feature \( V_f \).

\subsubsection{Multi-modal Feature Fusion}

To combine the features from both modalities, we fuse the weighted visual features \( V_f \) with the original textual features \( T \). The fusion can be performed either by concatenation or addition:
\begin{equation}
F_{\text{fused}} = \text{Concat}(V_f, T) \quad \text{or} \quad F_{\text{fused}} = V_f + T
\end{equation}
where \( F_{\text{fused}} \in \mathbb{R}^{2d} \) or \( \mathbb{R}^d \) is the fused multi-modal feature representation, which is used for downstream tasks such as stance-controlled generation.

\section{Experiments}
\begin{table*}[!t]
\centering
\begin{tabular}{cccccc}
\hline
\textbf{MODALITY}    & \textbf{MODEL}       & \textbf{Controllability ↑} & \textbf{CMSS ↑}            & \textbf{Relevance ↑}       &   \textbf{Perplexity ↓}               \\
\hline
Textual     & GPT4        & 0.8648          & 0.1951          & 0.5499          & 26.3243          \\
            & LLaMA3      & 0.8379          & 0.1985          & 0.5371          & \textbf{15.4041} \\
\hline
Visual      & GPT4-Vision & 0.7792          & 0.2175          & 0.5437          & 20.9887          \\
            & Qwen-VL     & 0.5764          & \underline{0.2674}    & \textbf{0.5463} & 19.2609          \\
\hline
Multi-modal & GPT4-Vision & \underline{0.9013}    & 0.2400          & 0.5098          & 22.5884          \\
            & Qwen-VL     & 0.6682          & \textbf{0.2825} & 0.4996          & \underline{17.5113}    \\
            & LLaVA       & 0.7214          & 0.2096          & 0.5173          & 198.5888         \\
            & LLaVA-SDMG       & \textbf{0.9257} & 0.1908          & \underline{0.5442}    & 58.6329    \\
\hline
\end{tabular}
\caption{Stance-driven controllable statement generation task performance on StanceGen2024, evaluating Relevance (↑), CMSS (↑), Controllability (↑), and Perplexity (↓). \textbf{Bold} indicates top performance; \underline{underline} marks second-best.}
\end{table*}

\begin{table*}[!t]
\centering
\resizebox{\textwidth}{!}{
\begin{tabular}{cccccccccc}
\hline
\multirow{2}{*}{\textbf{MODALITY}} & \multirow{2}{*}{\textbf{MODEL}} & \multicolumn{2}{c}{\textbf{Controllability ↑}} & \multicolumn{2}{c}{\textbf{CMSS ↑}} & \multicolumn{2}{c}{\textbf{Relevance ↑}} & \multicolumn{2}{c}{\textbf{Perplexity ↓}} \\
                                  &                                & \textbf{H} & \textbf{T} & \textbf{H} & \textbf{T} & \textbf{H} & \textbf{T} & \textbf{H} & \textbf{T} \\
\hline
\multirow{2}{*}{Textual}  & GPT4                           & 0.8515     & 0.8781     & 0.2105     & 0.1797     & 0.5661     & 0.5337     & 24.0868     & 28.5617     \\  
                                  & LLaMA3                         & 0.8511     & 0.8246     & 0.2102     & 0.1868     & 0.5477     & 0.5266     & \textbf{14.0936} & \textbf{16.7146} \\
\hline
\multirow{2}{*}{Visual}   & GPT4-Vision                    & 0.7427     & 0.8158     & 0.2225     & 0.2124     & 0.5487     & 0.5388     & 20.8939     & 21.0836     \\
                                  & Qwen-VL                        & 0.5369     & 0.6160     & \underline{0.2806} & \underline{0.2543} & \textbf{0.5517} & \textbf{0.5409} & 19.3511     & 19.1707     \\
\hline
\multirow{4}{*}{Multi-modal} & GPT4-Vision                 & \underline{0.8940} & \underline{0.9087} & 0.2404     & 0.2397     & 0.5177     & 0.5018     & 22.9812     & 22.1955     \\
                                       & Qwen-VL                      & 0.5777     & 0.7587     & \textbf{0.2889} & \textbf{0.2760} & 0.5067     & 0.4924     & \underline{17.8656} & \underline{17.1569} \\
                                       & LLaVA                         & 0.7386     & 0.7042     & 0.2168     & 0.2024     & 0.5180     & 0.5167     & 113.1138    & 284.0637    \\
                                       & LLaVA-SDMG                        & \textbf{0.9402} & \textbf{0.9112} & 0.1902     & 0.1915     & \underline{0.5489} & \underline{0.5395} & 54.7436     & 62.5221     \\
\hline
\end{tabular}
}
\caption{Stance-driven controllable statement generation Task Performance on StanceGen2024, evaluating Relevance (↑), CMSS (↑), Controllability (↑), and Perplexity (↓). The results are separated for Harris (H) and Trump (T) to highlight individual performance on each target. \textbf{Bold} indicates top performance; \underline{underline} marks second-best.}
\end{table*}

\subsection{Comparison Models}
~\textbf{Pure textual modality baselines:} 
(1) LLaMA3 ~\cite{dubey2024llama}, the Meta-Llama-3-8b-instruct; 
(2) GPT4 \footnote{https://openai.com/research/gpt-4}. \textbf{Multi-modal baselines:} 
(1) Qwen-VL~\cite{bai2023qwenvlversatilevisionlanguagemodel}, the Qwen-VL-Chat7b; 
(2) GPT4-Vision\footnote{https://openai.com/research/
gpt-4v-system-card}; 
(3) LLaVA~\cite{liu2023visual} (llava-v1.5-7b).

\subsection{Metrics}
To effectively evaluate the outcomes of our tasks, we employ the following metrics:
(1)~\textbf{Controllability:} Controllability measures the proportion of generated outputs that correctly exhibit the desired stance values, which is measured by employing a RoBERTa model ~\cite{liu2019robertarobustlyoptimizedbert} based classifier. 
(2)~\textbf{Perplexity:} Perplexity measures the fluency of replies, which is assessed by GPT-2 large.
(3)~\textbf{Relevance:} Relevance evaluates the contextual alignment between the real comments and the generated comments, calculated by the BAAI/bge-large-en-v1.5 model~\cite{xiao2024c}. 
(4)~\textbf{Cross-modal Semantic Similarity(CMSS):} Semantic Similarity evaluates how closely the generated text aligns with the content of the input image, calculated using the CLIP model~\cite{radford2021learning} (the clip-vit-large-patch14-336). This model computes the similarity in a shared embedding space for both text and image input.

\subsection{Instruction Finetuning}
We aim to enable the model to generate statements with a specific stance (favor or against) based on a given social media post, which includes both text and images. LLaVA’s vision-language understanding allows it to leverage both modalities, resulting in more contextually appropriate comments. LoRA fine-tuning enables the model to learn real social media commenting styles, making the generated text more natural.

We fine-tuned LLaVA with our SDMG Framework using DeepSpeed ZeRO-2~\cite{rajbhandari2020zero} and LoRA~\cite{hu2021lora}, resulting in the model referred to as LLaVA-SDMG. The dataset was split 8:2 for training and testing. Training used the AdamW optimizer with a learning rate of 2e-4, a batch size of 16, and a maximum sequence length of 2048 tokens.

\subsection{Result Analysis}

The performance of different LLMs and modality input on the Stance-driven controllable generation task for the StanceGen2024 dataset is respectively shown in Table 2 and Table 3.
\subsubsection{\textbf{Controllability} }
It can be seen that our proposed LLaVA-SDMG. demonstrates strong performance in controllability, particularly in the multi-modal setting, consistently outperforming its counterparts across different datasets.  For the Multi-modal task, LLaVA-SDMG achieves the highest controllability score with Harris (AVG: 0.9402) and Trump (AVG: 0.9112), indicating its superior ability to maintain control over the stance of generated content.  This outperforms other models, such as GPT4-Vision (AVG: 0.9013 for Harris and 0.9087 for Trump) and Qwen-VL (AVG: 0.6682 for Harris and 0.7587 for Trump), by a significant margin.

This is likely primarily due to its weighted multimodal feature fusion approach, as well as instruction fine-tuning.  The weighted fusion allows the model to flexibly adjust the importance of visual and textual information based on stance requirements during generation.  When the visual information strongly aligns with the stance, the model can increase the weight of visual features to enhance the influence of visual content on the generated text's stance, resulting in comments that better align with the stance requirements.  Additionally, instruction fine-tuning further improves the model's ability to understand and generate text that adheres to specific stance instructions, contributing to its strong stance controllability.

\subsubsection{\textbf{Response Quality}}
In terms of response quality, LLaVA-SDMG consistently demonstrates a strong balance between stance controllability and overall response quality.      
The relatively low correlation between generated text and images may stem from the weighted modality fusion process, where the model considers the input text to be more relevant to the stance and assigns it higher weight. As a result, the model focuses more on the stance rather than the image. Based on our observations, the images in the candidates' posts within our dataset predominantly convey the topic, with minimal impact on stance.


The relevance to real-world comments is second-best, while perplexity has improved significantly compared to the base model before enhancement, clearly resulting in better generation outcomes.
This is intuitive, as stance controllability and response quality can indeed be somewhat contradictory. It is difficult to ensure that generated sentences exhibit both strong stance controllability and high generation quality. Our approach effectively controls stance while preserving text fluency and relevance, demonstrating its ability to balance stance attribute preservation with maintaining the quality of the generated text.

Our approach is primarily applied to the open-source LLaVA model, and while it demonstrates some fluency weaknesses compared to powerful commercial large models, it still yields meaningful results.

\subsubsection{\textbf{Different Modal Inputs}}
The results in the table indicate that different modalities have varying impacts on the final outcomes.  Multimodal input significantly enhances the stance controllability of LLaVA-SDMG, but it also increases perplexity, suggesting challenges when handling complex multimodal tasks.  Overall, visual information has a limited impact on stance and mainly provides topic context.  Textual input plays a more significant role in stance controllability.  While multimodal input improves controllability, it may lead to a trade-off in the fluency and relevance of the generated text.  However, purely textual or visual input performs less effectively than multimodal input, as the latter results are more balanced and coherent.

\section{Conclusion}

This paper presents the new task of stance-driven multimodal controlled statement generation and introduces StanceGen2024, a novel dataset combining text, images, and video with stance annotations for political discourse. We propose a framework that integrates multimodal feature fusion with stance guidance, enhancing semantic consistency and stance control in generated textual statements. Our experiments show that the LLaVA-SDMG model, fine-tuned with this approach, effectively balances stance consistency with fluency. While challenges remain in fully leveraging visual content and ensuring fluency, our work lays the foundation for future research in  stance-controlled multimodal content generation.

\section*{Limitations}
The StanceGen2024 dataset focuses on the 2024 U.S. presidential election, limiting its generalizability to other political contexts or topics. Additionally, stance labeling in complex political discourse can be subjective, leading to potential inconsistencies despite efforts to ensure high-quality annotations.
Ethics Statement
\section*{Ethics Statement}
Political discourse is inherently biased, and stance detection may inadvertently amplify such biases.  The models trained on our dataset may reflect the political biases present in the original posts, and this could pose challenges for ensuring fairness and neutrality in generated content.
\bibliography{acl_latex}

\appendix

\end{document}